\documentclass{article}

\usepackage[numbers]{natbib}
\usepackage{mlncp_2024}
\usepackage{graphicx}
\usepackage{subcaption}
\usepackage{xcolor}
\usepackage{amsmath}  
\usepackage{amssymb}  
\usepackage{graphicx}

\usepackage{wrapfig}

\title{SLaNC: Static LayerNorm Calibration}
\author{Author Name \\ Affiliation}
\date{}

\begin{document}

\maketitle

\begin{abstract}
The ever increasing sizes of Large Language Models (LLMs) beyond hundreds of billions of parameters have generated enormous pressure on the manufacturers of dedicated hardware accelerators and made the innovative design of the latter one of the most rapidly expanding fields of the AI industry. Various approaches have been explored to enable efficient and accurate processing of LLMs on the available accelerators given their computational and storage limitations. Among these, various quantization techniques have become the main focus of the community as a means of reducing the compute, communication and storage requirements. Quantization to lower precision formats naturally poses a number of challenges caused by the limited range of the available value representations. When it comes to processing the popular Transformer models on hardware, one of the main issues becomes calculation of the LayerNorm simply because accumulation of the variance requires a much wider dynamic range than the hardware enables. In this article, we address this matter and propose a computationally-efficient scaling technique that can be easily applied to Transformer models during inference. Our method suggests a straightforward way of scaling the LayerNorm inputs based on the static weights of the immediately preceding linear layers. The scaling factors are computed offline, based solely on the linear layer weights, hence no latency or computational overhead is added during inference. Most importantly, our technique ensures that no numerical issues such as overflow or underflow could happen during the compute. This approach offers smooth, accurate and resource-effective inference across a wide range of hardware architectures. The article provides theoretical justification as well as supporting numerical simulations.
\end{abstract}

\section{Introduction}
Large Language Models (LLMs) based on Transformers \cite{vaswani2017attention} have recently become the dominant Deep Neural Network (DNN) architecture due to their unprecedented performance results in all language modeling \cite{Devlin2019BERTPO,Dai2019TransformerXLAL}, text processing \cite{Liu2019RoBERTaAR}, image and video generation \cite{dosovitskiy2021an}, and many other tasks. However, this success comes at a cost of enormous volumes of compute, storage, and data transfer. A whole new industry of dedicated hardware accelerators has emerged in the last few years to accommodate the needs of LLM training and inference \cite{gholami2022survey, wang2024beyond}. Another major initiative targeted at making the inference feasible and sustainable involves the development of lower precision formats \cite{NEURIPS2020_747e32ab, rouhani2023microscaling, soloveychik2022block}, efficient quantization techniques \cite{yao2022zeroquant}, algorithmic solutions \cite{frantar2023sparsegpt}, accurate approximations \cite{choromanski2020rethinking}, and other software optimizations \cite{wang2020linformer,qin2022cosformer}. 

Efficient quantization techniques such GPTQ \cite{frantar2022gptq}, AWQ \cite{lin2024awq}, SmoothQuant \cite{xiao2023smoothquant}, KVQuant \cite{hooper2024kvquant}, K-sort \cite{trukhanov2024accurate}, and numerous others enable storing and processing of LLMs in low-precision formats. Often, that would involve training the model in FP32 format and casting it to 4, 8 or 16-bit precision formats before deployment onto inference hardware \cite{yao2022zeroquant, darvish2023shared, trukhanov2024accurate}. The most popular approach is to compress the static weights to 4 or 8-bit integers or floats and reduce the activations to FP16 or BF16 \cite{micikevicius2017mixed}. In this paper, we focus on the wide family of accelerators operating on FP16 activations for their popularity \cite{9499865, jouppi2017datacenter} and specifically for the relatively narrow dynamic range (the range of representable numbers) of FP16 which might pose significant computational challenges. The most critical manifestation of this problem occurs during the LayerNorm computation. Importantly, inclusion of dozens or even hundreds of LayerNorm operators in current Transformers is unavoidable since they prevent the gradients from exploding or decaying during training \cite{brody2023expressivity}. At inference, though, processing LayerNorms on accelerators is extremely challenging because they require accumulation of squares of the inputs for the sake of variance (and norm) calculation \cite{guo2024slab}. Accumulation of such a large number of positive values in FP16 is almost surely bound to overflow.



In this work, we address this problem and propose an efficient, theoretically justified, and easy to implement scaling technique that leads to complete elimination of the FP16 overflow (or underflow) issue in LayerNorms. First, note that scaling of the LayerNorm input does not affect the output due to the homogeneity of the normalization operation but can very significantly shift the range of the accumulated numbers in the denominator computation. Based on this observation, we developed the SLaNC (Static LayerNorm Calibration) method which provides succinct closed formulae for scaling the inputs of all LayerNorms of any Transformer. Importantly, the SLaNC scales are computed solely based on the static weights of the preceding liner layers, and can be therefore computed offline without impacting the inference runtime. The formulae suggested by SLaNC are theoretically justified by derivations and detailed explanations and only involve norms of static weight matrices that can be directly and precisely computed using standard software. 

The rest of the article is organized as follows. First, we outline the notation, then in Section~\ref{prob_formula} we formulate the numerical problem caused by the LayerNorm computation in FP16. Section~\ref{slanc} presents the SLaNC technique together with its theoretical justification. Supporting numerical simulation on the Llama family of LLMs are demonstrated in Section~\ref{num_analysis}. The concluding remarks can be found in Section~\ref{conclusion}.
 
\noindent \textbf{Notation.}
The following notation is used in the article. Matrices are denoted by capital bold letters $\mathbf{M}$ and vectors by lower case bold $\mathbf{v}$. The operator product of matrices $\mathbf{A}$ and $\mathbf{B}$ of appropriate sizes is written as $\mathbf{A}\cdot\mathbf{B}$ or $\mathbf{A}\mathbf{B}$, while their element-wise product would be denoted by $\mathbf{A}\odot\mathbf{B}$. For matrix $\mathbf{M}$, we write $\| \mathbf{M} \|_F$ for its Frobenius norm and $\| \mathbf{M} \|$ for its spectral norm; for vector $\mathbf{v}$, by $\|\mathbf{v}\|$ we denote its Euclidean norm. Given vector $\mathbf{m}$, we denote by $\mathbf{M} = \text{diag}(\mathbf{m})$ the diagonal matrix with elements of $\mathbf{m}$ on the main diagonal.

\section {Problem Formulation}
\label{prob_formula}
Quantization of an LLM to a low-precision format (e.g., 4, 8 or 16-bit) can lead to a significant degradation of the output quality, and thus has to be applied together with some advanced technique capable of restoring the accuracy \cite{frantar2022gptq, lin2024awq, xiao2023smoothquant,hooper2024kvquant, trukhanov2024accurate, Bondarenko2021UnderstandingAO, Nagel2021AWP}. However, an even bigger challenge caused by casting models into low-precision formats is the limited dynamic range of such formats, which can completely ruin the compute flow if applied blindly. The most prominent example is the computation of LayerNorm, which becomes impossible on FP16 accelerators due to the unavoidable overflows and underflows as demonstrated next.

\subsection{LayerNorm Compute}
\label{prob_formul}

Layer Normalization (LayerNorm) has become one of the most ubiquitous non-linear operations in modern DNNs since it prevents the gradients from decaying or exploding during training. Extensive literature has demonstrated that the current DNN architectures cannot be practically trained without frequent normalization of hidden states \cite{ba2016layer, batchnormal, wu2018group}. State of the art Transformer models include dozens or even hundreds of LayerNorm operators which are introduced to facilitate training but make inference troublesome due to the numerical problems introduced by the computation of their denominators.

Given a row input $\mathbf{x} \in \mathbb{R}^d$ and fixed parameters $\boldsymbol{\gamma},\;\boldsymbol{\beta} \in \mathbb{R}^d$, the LayerNorm output reads as 
\begin{equation}
\label{eq_laynorm}
 \mathbf{y}(\mathbf{x}) = \left(\frac{\mathbf{x}-{\mu}\mathbf{1}}{\sigma}\right) * \boldsymbol{\gamma}+\boldsymbol{\beta} = \left(\frac{\mathbf{x}-{\mu}\mathbf{1}}{\sigma}\right) \mathbf{\Gamma}+\boldsymbol{\beta} \in \mathbb{R}^d, 
 \end{equation}
where $\mathbf{1} \in \mathbb{R}^d$ is the vector of ones, $\mathbf{\Gamma} = \text{diag}(\boldsymbol{\gamma})$, and
\begin{equation}
\label{eq:var_def}
\mathbf{\mu} = \frac{1}{d} \sum_{i=1}^d x_i, ~~~\text{and} ~~~ \sigma = \sqrt{\frac{1}{d} \sum_{i=1}^d (x_i-\mu)^2} = \sqrt{\frac{1}{d} \sum_{i=1}^d x_i^2 - \mu^2}.
\end{equation}

\begin{wrapfigure}{r}{0.38\textwidth}
\vspace{-0.5cm}
    \centering
    \begin{minipage}[b]{0.4\textwidth}
        \centering
\begin{minipage}[b]{0.43\textwidth}
        \centering
        \includegraphics[width=\linewidth]{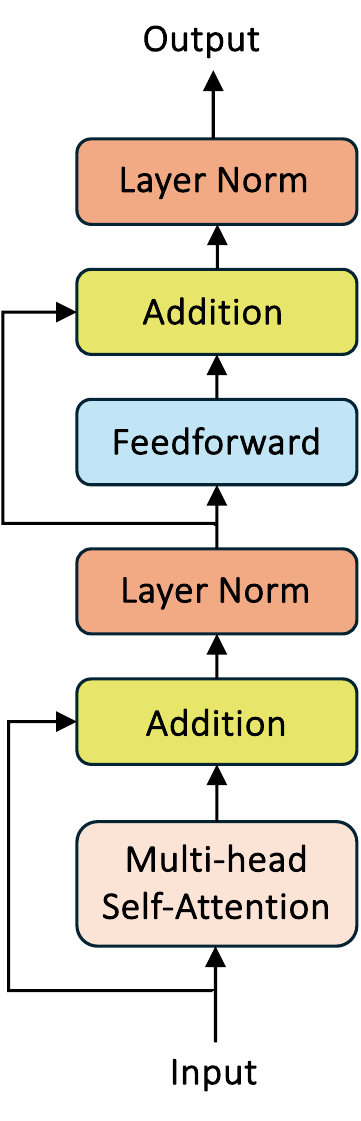}
        \subcaption{Post-LN}
        \label{fig_arch1}
    \end{minipage}\hfill
    \begin{minipage}[b]{0.41 \textwidth}
        \centering
        \includegraphics[width=\linewidth]{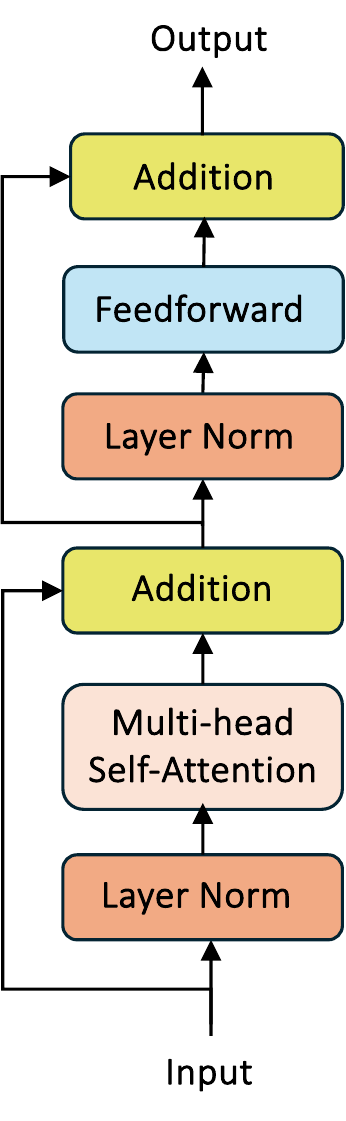}
        \subcaption{Pre-LN}
        \label{fig_arch2}   
    \end{minipage}
    \caption{Residual branching options.}
    \label{fig_arch_var} 
    \end{minipage}
    \end{wrapfigure}

As Eq.~\ref{eq:var_def} suggests, the standard way of computing $\sigma$ requires summing up the squares of the input vector elements. Depending on the range of these elements, such accumulation can easily lead to an overflow or underflow when performed in FP16 or FP8 formats. It is important to note that the majority of the available LLM accelerators process non-linear operations exclusively in FP16 format \cite{markidis2018nvidia, mudigere2022software}. While some accelerators do support FP32 accumulation in non-linear modules, this option often comes at a high latency increase making FP32 regime impractical.
Fig.~\ref{sq_sum:subfig1} and Fig.~\ref{sq_sum:subfig3} show the typical distributions of the sum of squares from Eq.~\ref{eq:var_def} in one of the layers of Llama-2. We observe that in too many cases the resulting values exceed the range of FP16, leading to invalid inference.

Note also that the Transformer architecture comes in two flavors based on the location of the residual branch-out. It can take off before the LayerNorm (pre-LN residual) or after (post-LN residual), Fig.~\ref{fig_arch_var}. Originally, the post-LN option was suggested  \cite{vaswani2017attention} but later the other one became quite popular since it was observed to speeds-up the training \cite{xiong2020layer}. To be specific and for lack of space below we focus on the post-LN desugn, however, we emphasize that the derivations and conclusions equally apply to the pre-LN one.

\section{LayerNorm Scaling}
\label{slanc}

\subsection{Dynamic Model Calibration}
The natural way of addressing the problem of overflow or underflow during computation of LayerNorm would be to appropriately scale its input. Determining the correct scaling factors appears to be challenging because while avoiding overflow we also do not want to excessively dump the input causing underflow and vice versa. As a consequence, any reasonable scaling algorithm must take into account the actual LayerNorm input values and cannot set the scaling parameters blindly. 

A common solution would be to calibrate the scaling factors. This involves passing a test dataset through the Transformer to gauge the range of the input vector norms and setting the scaling factor based on some first-order statistic of this range (e.g., mean or median norm). This technique requires extra calibration data and significant computational overhead even for such a basic operation as LayerNorm, making this approach impractical.

\subsection{Analytical Static Scaling}
In this work, we propose a different methodology that enables analytical offline computation of the desired scaling factors. The scales are determined solely based on the static weights of the liner layers immediately preceding the LayerNorm at hand. This way we calibrate all the LayerNorm operators of a model statically, without using a calibration dataset or additional runtime resources — everything is computed preemptively during model compilation.

The idea of the method is based on a simple observation that LayerNorms inside a Transformer occur frequently and in a regular pattern since any large Transformer is a chain of dozens of identical decoders. Typically, two consecutive LayerNorms surround the attention or the Multi-Layer Perceptron (MLP) block of every decoder. Eq.~\ref{eq_laynorm} suggests that we can treat a LayerNorm as a Euclidean normalization followed by a diagonal matrix multiplication.\footnote{Since we are mainly focusing on the order of magnitude of the norms of the hidden states involved, without impact on accuracy we discard the additive biases $\boldsymbol{\beta}$ of the LayerNorm operator.} From this natural decomposition of the LayerNorm operator we infer that immediately after normalization (the first step in LayerNorm), the norm of the hidden vector $\mathbf{x}$ is equal to one. Our goal is to trace the computational graph from this point to the next LayerNorm and gauge the orders of magnitude of $\|\mathbf{x}\|$ changes based on the transformations it undergoes along the way.

\subsection{SLaNC for Standard MLP Block}
\label{sec_mlp}
\begin{figure}[h]
    \centering
    \begin{subfigure}[b]{0.21\textwidth}
        \centering
        \includegraphics[width=\textwidth]{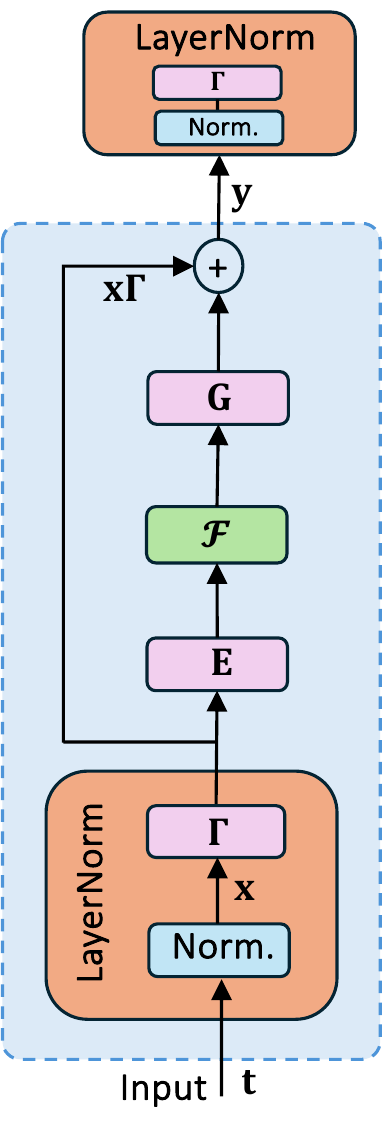}
        \subcaption{Standard MLP}
        \label{fig_standrad_mlp}
    \end{subfigure}
    \hfill
    \begin{subfigure}[b]{0.25\textwidth}
        \centering
        \includegraphics[width=\textwidth]{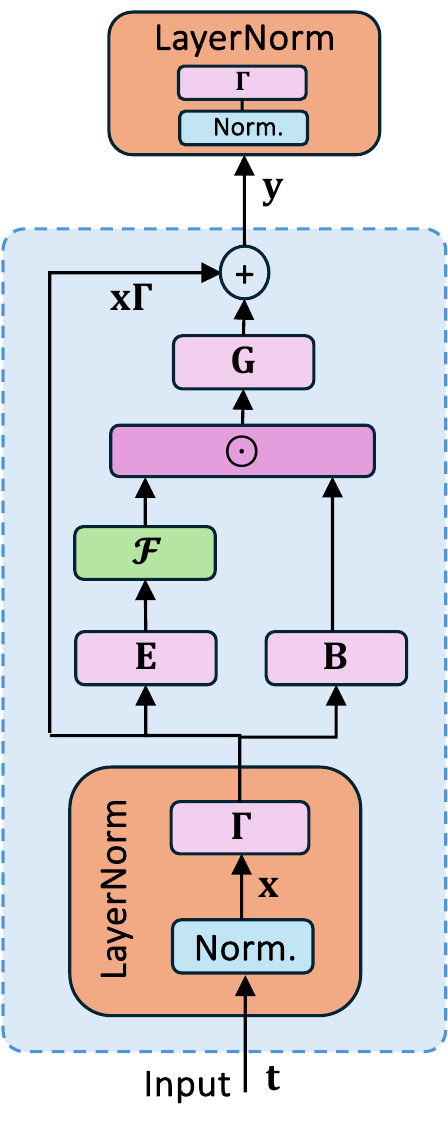}
        \subcaption{Llama MLP}
        \label{fig_llama_mlp}
    \end{subfigure}   
     \hfill
        \begin{subfigure}[b]{0.32\textwidth}
        \centering
        \includegraphics[width=\textwidth]{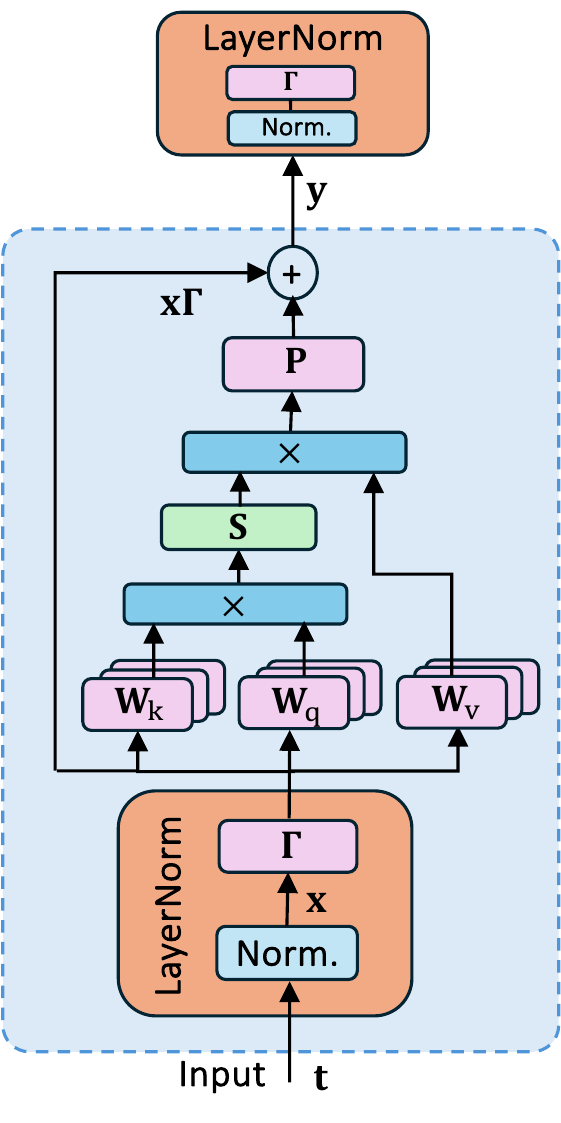}
        \subcaption{Attention block}
        \label{fig_attn}
    \end{subfigure}

    \caption{The compute flow between consecutive LayerNorms of various Transformers. Pink blocks with capital letters stand for linear layers with the corresponding weight matrices, green $\mathcal{F}$-blocks represent non-linearities and $\mathbf{S}$-block represents $\operatorname{Softmax}$.}
    \label{fig_transformer_archs}
\end{figure}

To illustrate the idea, let us consider the MLP block of a standard Transformer, Fig.~\ref{fig_standrad_mlp}. Since we neglect the additive bias $\boldsymbol{\beta}$, the output of the MLP block can be expressed as
\begin{equation}
\mathbf{y} = \mathcal{F} \left(\mathbf{x} \mathbf{\Gamma} \mathbf{E}\right)\mathbf{G} + \mathbf{x} \mathbf{\Gamma},
\end{equation}
where the addition comes from the residual connection, and $\mathcal{F}(\cdot)$ is an element-wise non-linearity which is usually a contraction function (e.g. ReLU, GeLU, etc.) making the norm of its argument smaller. Since usually, the maximal partial derivative of $\mathcal{F}(\cdot)$ is bounded by a constant close to one, we can approximate the norm of $\mathbf{y}$ as
\begin{equation}
\|\mathbf{y}\| \propto \|\mathbf{x} \mathbf{\Gamma} \mathbf{E}\mathbf{G} + \mathbf{x} \mathbf{\Gamma}\|_F.
\end{equation}
Eventually, we conclude that
\begin{equation}
\label{scale_factor}
\frac{\|\mathbf{y}\|}{\|\mathbf{x}\|} \propto \|\mathbf{\Gamma} (\mathbf{E}\mathbf{G} + \mathbf{I})\|_F.
\end{equation}
Recall that $\mathbf{x}$ is the output of the normalization step of a LayerNorm (see Fig. \ref{fig_standrad_mlp}) and thus has unit norm. Therefore, it is natural to set the scaling factor of the following LayerNorm to the right-hand side of Eq.~\ref{scale_factor} and this should solve the overflow/underflow issue. In Section~\ref{num_analysis}, we demonstrate by extensive simulations that this is actually the case. Note that the scale determined by Eq.~\ref{scale_factor} only involves static weights and can be computed offline. 

\subsection{SLaNC for Llama MLP Block}
Using the same methodology, we derive an analogous formula for the scaling factors of the LayerNorm following the modified MLP block designed for the decoders of the Llama family of models, Fig.~\ref{fig_llama_mlp}. Here, in addition to the two linear layers of the standard MLP block, we have another linear layer whose output is multiplied with the output of the non-linearity in the element-wise manner. The non-liner function itself is usually chosen to be GeLU. The input of the post-MLP block LayerNorm $\mathbf{y}$ reads as
\begin{equation}
\mathbf{y} = \left( \mathcal{F} \left(\mathbf{x} \mathbf{\Gamma} \mathbf{E}\right) \odot \mathbf{x} \mathbf{\Gamma}\mathbf{B}\right)\mathbf{G} + \mathbf{x} \mathbf{\Gamma}.
\end{equation}
Similar principles as above together with basic properties of matrix norms yield
\begin{equation}
\|\mathbf{y}\| \propto \|\|\mathbf{\Gamma} \mathbf{E}\| \mathbf{x} \mathbf{\Gamma}\mathbf{B}\mathbf{G} + \mathbf{x} \mathbf{\Gamma}\|_F,
\end{equation}
where we used the fact that $\|\mathbf{x}\| = 1$. Finally, the scaling factor computes as 
\begin{equation}
\frac{\|\mathbf{y}\|}{\|\mathbf{x}\|} \propto \|\mathbf{\Gamma}\left(\|\mathbf{\Gamma} \mathbf{E}\|\mathbf{B}\mathbf{G} + \mathbf{I}\right)\|_F.
\end{equation}

\subsection{SLaNC for the Attention Block}
Next, we derive a formula for the scaling factor of the LayerNorm following the standard attention block with $h$ heads. As it can be seen in Fig.~\ref{fig_attn}, the most critical observation here is that the product of the $\operatorname{Softmax}$ output $\mathbf{S}^i$ of head $i$ with $\mathbf{V}^i$ results in a convex combination of the rows of the latter. The outputs $\{\mathbf{S}^i\mathbf{V}^i\}_{i=1}^h$ are concatenated, hence, the norm of the concatenated vector can be approximated by the norm of the concatenation of $\{\mathbf{x}\mathbf{\Gamma}\mathbf{W}_{\mathbf{V}}^i\}_{i=1}^h$ which is precisely $\mathbf{x}\mathbf{\Gamma}\mathbf{W}_{\mathbf{V}}$. We get
\begin{equation}
\|\mathbf{y}\| \propto \|\mathbf{x}\mathbf{\Gamma}\mathbf{W}_{\mathbf{V}}\mathbf{P} + \mathbf{x}\mathbf{\Gamma}\|_F = \|\mathbf{x}\mathbf{\Gamma}(\mathbf{W}_{\mathbf{V}}\mathbf{P} + \mathbf{I})
\|_F,
\end{equation}
and conclude that the following scale should be used in the post-attention LayerNorm operator
\begin{equation}
\frac{\|\mathbf{y}\|}{\|\mathbf{x}\|} \propto \|\mathbf{\Gamma}(\mathbf{W}_{\mathbf{V}}\mathbf{P} + \mathbf{I})
\|_F.
\end{equation}

\begin{figure}[h]
    \centering

    \begin{subfigure}[b]{0.8\textwidth}
        \centering
        \includegraphics[width=\textwidth]{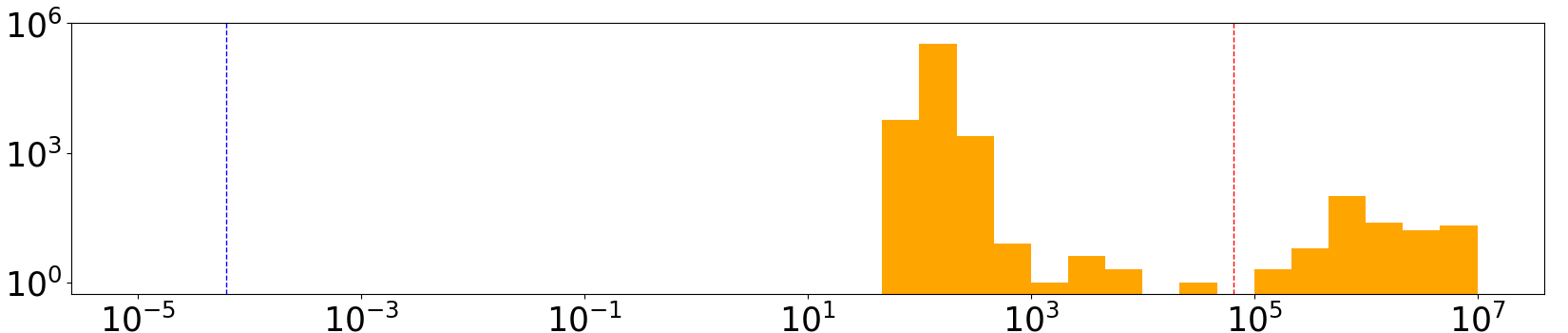} 
        \caption{post-attention RMSNorm input, unscaled}
        \label{sq_sum:subfig1}
    \end{subfigure}
    
    \vspace{0.1cm} 

    \begin{subfigure}[b]{0.8\textwidth}
        \centering
        \includegraphics[width=\textwidth]{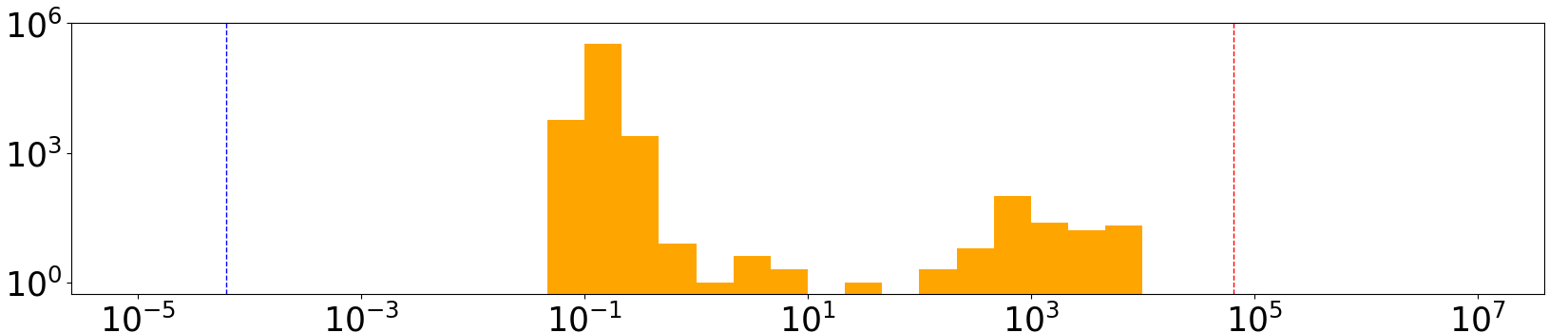} 
        \caption{post-attention RMSNorm input, SLaNC scaled}
        \label{sq_sum:subfig2}
    \end{subfigure}

    \vspace{0.1cm} 

    \begin{subfigure}[b]{0.8\textwidth}
        \centering
        \includegraphics[width=\textwidth]{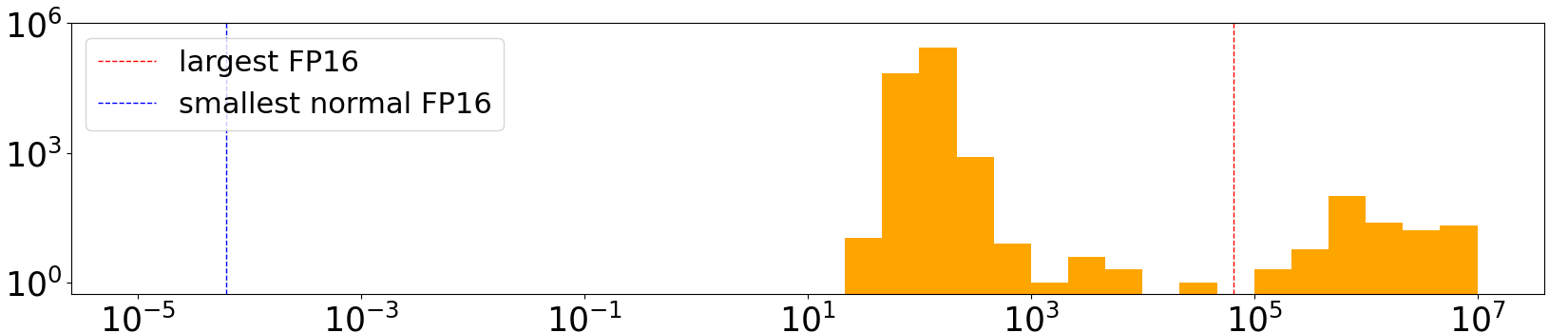} 
        
        \caption{post-MLP RMSNorm, unscaled}
        \label{sq_sum:subfig3}
    \end{subfigure}
    
    \vspace{0.1cm} 
    
    \begin{subfigure}[b]{0.8\textwidth}
        \centering
        \includegraphics[width=\textwidth]{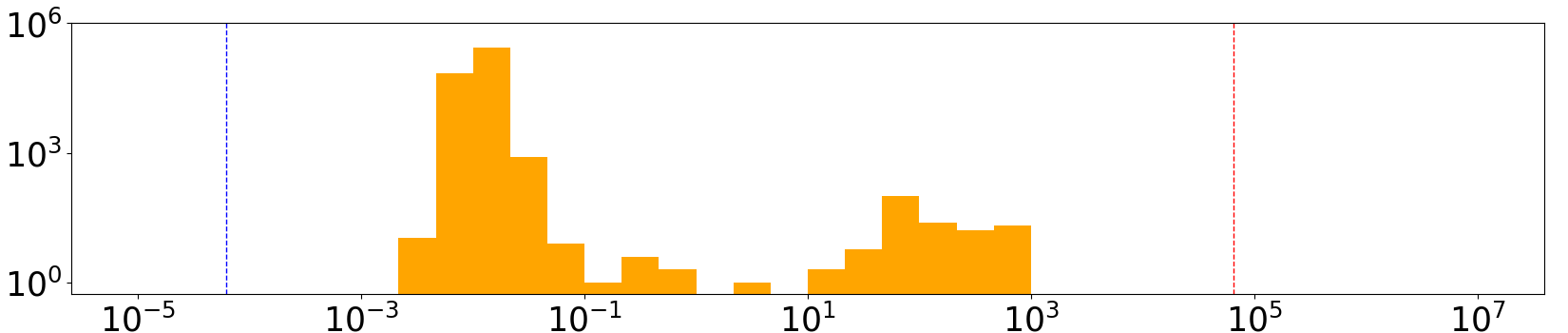} 
        \caption{post-MLP RMSNorm, SLaNC scaled}
        \label{sq_sum:subfig4}
    \end{subfigure}
    
    \caption{Empirical histograms of the sum of squares in RMSNorm layers of the 9th decoder in Llama-2-7b, calculated on wikitext2. The red vertical cut-off line sets the maximal representable FP16 value (65k) beyond which FP16 overflows, the blue line shows the minimal normal FP16 value. Histograms (b) and (d) show that after SLaNC scaling no overflow (or underflow) is detected and the RMSNorm is computed precisely.}
    \label{sq_sum}
\end{figure}

\section{Experiments} 
\label{num_analysis}
To demonstrate the power of our SLaNC scaling technique, we present simulation results for Llama models. Note that the Llama architecture replaces the LayerNorm by Root Mean Squared Layer Normalization (RMSNorm) \cite{zhang2019rootmeansquarelayer}, which differs from the former only by omitting the mean $\mu$ subtraction in Eq.~\ref{eq:var_def} and thus does not affect SLaNC scaling. 

In our first experiment, we collected empirical statistics of the sums of squares in the denominators of the RMSNorm operators without scaling and with SLaNC scaling. To this end, we applied Llama-2-7b to Wikitext~2 dataset. Fig.~\ref{sq_sum:subfig1} and \ref{sq_sum:subfig3} feature typical histograms in two consecutive RMSNorms of this model. We see that in a significant number of cases, the sum of squares well exceeds the FP16 range and causes overflow. The SLaNC scaling changes the situation dramatically and not only shifts the histograms inside the FP16 range but also keeps safe margins on both edges of the range, as illustrated by Fig.~\ref{sq_sum:subfig2} and \ref{sq_sum:subfig4}, respectively.

Next, we compared the perplexities of Llama models on the same Wikitext~2 dataset with the default FP32 implementation of RMSNorm and with the sum of squared accumulated in FP16 (all other operations from the default setup intact). Table~\ref{LLama wikitext perplexity} shows a significant degradation when the accumulation happens in FP16 exactly due to numerous overflows. This problem is completely resolved when the SLaNC scaling is applied. We also note that in all standard models, a small constant $\varepsilon$ is added to the variance of the input in the denominator of LayerNorm or RMSNorm operator. This way we can avoid division by zero in the case of underflow and improve the numerical stability. Since SLaNC scales are known ahead of time, we can easily apply them to the $\varepsilon$ constants as well (in fact, we divide $\varepsilon$ by the squared SLaNC scalings). As the bottom row of Table~\ref{LLama wikitext perplexity} demonstrates, now the FP16 SLaNC scaling can precisely reproduce the default FP32 values.



\begin{table}[h!]
\centering
\caption{Llama perplexity on Wikitext~2 with different RMSNorm computation modes.}
 \vspace{8pt}
\begin{tabular}{|c|c|c|c|}
\hline
 accumulation format & Llama-2-7b & Llama-2-13b & Llama-3-8b\\
\hline
FP32 & 5.116 & 4.574 & 5.538 \\
FP16  & 19.105 & 10.521 & 16.013 \\
FP16 + SLaNC  & 5.116 & 4.573 & 5.539\\
\hline
\end{tabular}
\label{LLama wikitext perplexity}
\end{table}

\section{Conclusion}
\label{conclusion}
In this paper, we present a novel SLaNC technique that makes 
LLM inference possible on FP16 accelerators without the need to cast LayerNorm operators into FP32. This theoretically grounded approach provides easy-to-use formulae for an offline computation of scaling factors for the inputs of LayerNorms. The SLaNC scaling factors guarantee precise computation of the LayerNorm in FP16 and provably avoid overflows and underflows. By keeping all the compute in FP16, the SLaNC algorithm enables low latency accurate compute, which is demonstrated by our extensive numerical simulations.


\bibliographystyle{unsrt}
\setcounter{NAT@ctr}{0}
\bibliography{references}

\begin{thebibliography}{10}

\bibitem{vaswani2017attention}
Ashish Vaswani, Noam Shazeer, Niki Parmar, Jakob Uszkoreit, Llion Jones,
  Aidan~N Gomez, {\L}ukasz Kaiser, and Illia Polosukhin.
\newblock Attention is all you need.
\newblock In {\em Advances in Neural Information Processing Systems}, pages
  5998--6008, 2017.

\bibitem{Devlin2019BERTPO}
Jacob Devlin, Ming-Wei Chang, Kenton Lee, and Kristina Toutanova.
\newblock {BERT}: Pre-training of deep bidirectional {T}ransformers for
  language understanding.
\newblock In {\em North American Chapter of the Association for Computational
  Linguistics}, 2019.

\bibitem{Dai2019TransformerXLAL}
Zihang Dai, Zhilin Yang, Yiming Yang, Jaime~G. Carbonell, Quoc~V. Le, and
  Ruslan Salakhutdinov.
\newblock Transformer-{XL}: Attentive language models beyond a fixed-length
  context.
\newblock In {\em Annual Meeting of the Association for Computational
  Linguistics}, 2019.

\bibitem{Liu2019RoBERTaAR}
Yinhan Liu, Myle Ott, Naman Goyal, Jingfei Du, Mandar Joshi, Danqi Chen, Omer
  Levy, Mike Lewis, Luke Zettlemoyer, and Veselin Stoyanov.
\newblock Ro{BERT}a: A robustly optimized {BERT} pretraining approach.
\newblock {\em arXiv preprint arXiv:1907.11692}, 2019.

\bibitem{dosovitskiy2021an}
Alexey Dosovitskiy, Lucas Beyer, Alexander Kolesnikov, Dirk Weissenborn,
  Xiaohua Zhai, Thomas Unterthiner, Mostafa Dehghani, Matthias Minderer, Georg
  Heigold, Sylvain Gelly, Jakob Uszkoreit, and Neil Houlsby.
\newblock An image is worth 16x16 words: {T}ransformers for image recognition
  at scale.
\newblock In {\em International Conference on Learning Representations}, 2021.

\bibitem{gholami2022survey}
Amir Gholami, Sehoon Kim, Zhen Dong, Zhewei Yao, Michael~W Mahoney, and Kurt
  Keutzer.
\newblock A survey of quantization methods for efficient neural network
  inference.
\newblock In {\em Low-Power Computer Vision}, pages 291--326. Chapman and
  Hall/CRC, 2022.

\bibitem{wang2024beyond}
Xindi Wang, Mahsa Salmani, Parsa Omidi, Xiangyu Ren, Mehdi Rezagholizadeh, and
  Armaghan Eshaghi.
\newblock Beyond the limits: A survey of techniques to extend the context
  length in large language models.
\newblock {\em arXiv preprint arXiv:2402.02244}, 2024.

\bibitem{NEURIPS2020_747e32ab}
Bita Darvish~Rouhani, Daniel Lo, Ritchie Zhao, Ming Liu, Jeremy Fowers, Kalin
  Ovtcharov, Anna Vinogradsky, Sarah Massengill, Lita Yang, Ray Bittner,
  Alessandro Forin, Haishan Zhu, Taesik Na, Prerak Patel, Shuai Che, Lok
  Chand~Koppaka, XIA SONG, Subhojit Som, Kaustav Das, Saurabh T, Steve
  Reinhardt, Sitaram Lanka, Eric Chung, and Doug Burger.
\newblock Pushing the limits of narrow precision inferencing at cloud scale
  with {M}icrosoft floating point.
\newblock In H.~Larochelle, M.~Ranzato, R.~Hadsell, M.F. Balcan, and H.~Lin,
  editors, {\em Advances in Neural Information Processing Systems}, volume~33,
  pages 10271--10281. Curran Associates, Inc., 2020.

\bibitem{rouhani2023microscaling}
Bita~Darvish Rouhani, Ritchie Zhao, Ankit More, Mathew Hall, Alireza
  Khodamoradi, Summer Deng, Dhruv Choudhary, Marius Cornea, Eric Dellinger,
  Kristof Denolf, et~al.
\newblock Microscaling data formats for deep learning.
\newblock {\em arXiv preprint arXiv:2310.10537}, 2023.

\bibitem{soloveychik2022block}
Ilya Soloveychik, Ilya Lyubomirsky, Xin Wang, and Sudeep Bhoja.
\newblock Block format error bounds and optimal block size selection.
\newblock {\em AAAI Conference, ENC2 workshop}, 2023.

\bibitem{yao2022zeroquant}
Zhewei Yao, Reza Yazdani~Aminabadi, Minjia Zhang, Xiaoxia Wu, Conglong Li, and
  Yuxiong He.
\newblock Zero{Q}uant: Efficient and affordable post-training quantization for
  large-scale {T}ransformers.
\newblock {\em Advances in Neural Information Processing Systems},
  35:27168--27183, 2022.

\bibitem{frantar2023sparsegpt}
Elias Frantar and Dan Alistarh.
\newblock Sparse{GPT}: Massive language models can be accurately pruned in
  one-shot.
\newblock In {\em International Conference on Machine Learning}, pages
  10323--10337. PMLR, 2023.

\bibitem{choromanski2020rethinking}
Krzysztof Choromanski, Valerii Likhosherstov, David Dohan, Xingyou Song,
  Andreea Gane, Tamas Sarlos, Peter Hawkins, Jared Davis, Afroz Mohiuddin,
  Lukasz Kaiser, et~al.
\newblock Rethinking attention with performers.
\newblock {\em arXiv preprint arXiv:2009.14794}, 2020.

\bibitem{wang2020linformer}
Sinong Wang, Belinda~Z Li, Madian Khabsa, Han Fang, and Hao Ma.
\newblock Linformer: Self-attention with linear complexity.
\newblock {\em arXiv preprint arXiv:2006.04768}, 2020.

\bibitem{qin2022cosformer}
Zhen Qin, Weixuan Sun, Hui Deng, Dongxu Li, Yunshen Wei, Baohong Lv, Junjie
  Yan, Lingpeng Kong, and Yiran Zhong.
\newblock cosformer: Rethinking softmax in attention.
\newblock {\em arXiv preprint arXiv:2202.08791}, 2022.

\bibitem{frantar2022gptq}
Elias Frantar, Saleh Ashkboos, Torsten Hoefler, and Dan Alistarh.
\newblock {GPTQ}: Accurate post-training quantization for generative
  pre-trained {T}ransformers.
\newblock {\em arXiv preprint arXiv:2210.17323}, 2022.

\bibitem{lin2024awq}
Ji~Lin, Jiaming Tang, Haotian Tang, Shang Yang, Wei-Ming Chen, Wei-Chen Wang,
  Guangxuan Xiao, Xingyu Dang, Chuang Gan, and Song Han.
\newblock {AWQ}: Activation-aware weight quantization for on-device {LLM}
  compression and acceleration.
\newblock {\em Proceedings of Machine Learning and Systems}, 6:87--100, 2024.

\bibitem{xiao2023smoothquant}
Guangxuan Xiao, Ji~Lin, Mickael Seznec, Hao Wu, Julien Demouth, and Song Han.
\newblock Smooth{Q}uant: Accurate and efficient post-training quantization for
  large language models.
\newblock In {\em International Conference on Machine Learning}, pages
  38087--38099. PMLR, 2023.

\bibitem{hooper2024kvquant}
Coleman Hooper, Sehoon Kim, Hiva Mohammadzadeh, Michael~W Mahoney, Yakun~Sophia
  Shao, Kurt Keutzer, and Amir Gholami.
\newblock {KVQ}uant: Towards 10 million context length {LLM} inference with
  {KV} cache quantization.
\newblock {\em arXiv preprint arXiv:2401.18079}, 2024.

\bibitem{trukhanov2024accurate}
Nikita Trukhanov and Ilya Soloveychik.
\newblock Accurate block quantization in {LLM}s with outliers.
\newblock {\em AAAI Conference, ENC2 {W}orkshop}, 2024.

\bibitem{darvish2023shared}
Bita Darvish~Rouhani, Ritchie Zhao, Venmugil Elango, Rasoul Shafipour, Mathew
  Hall, Maral Mesmakhosroshahi, Ankit More, Levi Melnick, Maximilian Golub,
  Girish Varatkar, et~al.
\newblock With shared microexponents, a little shifting goes a long way.
\newblock In {\em Proceedings of the 50th Annual International Symposium on
  Computer Architecture}, pages 1--13, 2023.

\bibitem{micikevicius2017mixed}
Paulius Micikevicius, Sharan Narang, Jonah Alben, Gregory Diamos, Erich Elsen,
  David Garcia, Boris Ginsburg, Michael Houston, Oleksii Kuchaiev, Ganesh
  Venkatesh, and Hao Wu.
\newblock Mixed precision training.
\newblock In {\em International Conference on Learning Representations}, 2018.

\bibitem{9499865}
Swagath Venkataramani, Vijayalakshmi Srinivasan, Wei Wang, Sanchari Sen, Jintao
  Zhang, Ankur Agrawal, Monodeep Kar, Shubham Jain, Alberto Mannari, Hoang
  Tran, Yulong Li, Eri Ogawa, Kazuaki Ishizaki, Hiroshi Inoue, Marcel Schaal,
  Mauricio Serrano, Jungwook Choi, Xiao Sun, Naigang Wang, Chia-Yu Chen,
  Allison Allain, James Bonano, Nianzheng Cao, Robert Casatuta, Matthew Cohen,
  Bruce Fleischer, Michael Guillorn, Howard Haynie, Jinwook Jung, Mingu Kang,
  Kyu-hyoun Kim, Siyu Koswatta, Saekyu Lee, Martin Lutz, Silvia Mueller,
  Jinwook Oh, Ashish Ranjan, Zhibin Ren, Scot Rider, Kerstin Schelm, Michael
  Scheuermann, Joel Silberman, Jie Yang, Vidhi Zalani, Xin Zhang, Ching Zhou,
  Matt Ziegler, Vinay Shah, Moriyoshi Ohara, Pong-Fei Lu, Brian Curran, Sunil
  Shukla, Leland Chang, and Kailash Gopalakrishnan.
\newblock Ra{P}i{D}: {AI} accelerator for ultra-low precision training and
  inference.
\newblock In {\em 2021 ACM/IEEE 48th Annual International Symposium on Computer
  Architecture (ISCA)}, pages 153--166, 2021.

\bibitem{jouppi2017datacenter}
Norman~P Jouppi, Cliff Young, Nishant Patil, David Patterson, Gaurav Agrawal,
  Raminder Bajwa, Sarah Bates, Suresh Bhatia, Nan Boden, Al~Borchers, et~al.
\newblock In-datacenter performance analysis of a tensor processing unit.
\newblock In {\em Proceedings of the 44th Annual International Symposium on
  Computer Architecture}, pages 1--12, 2017.

\bibitem{brody2023expressivity}
Shaked Brody, Uri Alon, and Eran Yahav.
\newblock On the expressivity role of layernorm in {T}ransformers' attention.
\newblock In {\em Annual Meeting of the Association for Computational
  Linguistics}, 2023.

\bibitem{guo2024slab}
Jialong Guo, Xinghao Chen, Yehui Tang, and Yunhe Wang.
\newblock {SLAB}: Efficient {T}ransformers with simplified linear attention and
  progressive re-parameterized batch normalization.
\newblock {\em arXiv preprint arXiv:2405.11582}, 2024.

\bibitem{Bondarenko2021UnderstandingAO}
Yelysei Bondarenko, Markus Nagel, and Tijmen Blankevoort.
\newblock Understanding and overcoming the challenges of efficient
  {T}ransformer quantization.
\newblock {\em ArXiv}, abs/2109.12948, 2021.

\bibitem{Nagel2021AWP}
Markus Nagel, Marios Fournarakis, Rana~Ali Amjad, Yelysei Bondarenko, Mart
  Van~Baalen, and Tijmen Blankevoort.
\newblock A white paper on neural network quantization.
\newblock {\em arXiv preprint arXiv:2106.08295}, 2021.

\bibitem{ba2016layer}
Jimmy Ba, Jamie~Ryan Kiros, and Geoffrey~E Hinton.
\newblock Layer normalization.
\newblock {\em arXiv preprint arXiv:1607.06450}, 2016.

\bibitem{batchnormal}
Sergey Ioffe and Christian Szegedy.
\newblock Batch normalization: Accelerating deep network training by reducing
  internal covariate shift.
\newblock In {\em Proceedings of the 32nd International Conference on
  International Conference on Machine Learning - Volume 37}, ICML'15, page
  448–456. JMLR.org, 2015.

\bibitem{wu2018group}
Yuxin Wu and Kaiming He.
\newblock Group normalization.
\newblock In {\em Proceedings of the European Conference on Computer Vision
  (ECCV)}, pages 3--19, 2018.

\bibitem{markidis2018nvidia}
Stefano Markidis, Steven~Wei Der~Chien, Erwin Laure, Ivy~Bo Peng, and Jeffrey~S
  Vetter.
\newblock {NVIDIA} tensor core programmability, performance \& precision.
\newblock In {\em 2018 IEEE International Parallel and Distributed Processing
  Symposium Workshops (IPDPSW)}, pages 522--531. IEEE, 2018.

\bibitem{mudigere2022software}
Dheevatsa Mudigere, Yuchen Hao, Jianyu Huang, Zhihao Jia, Andrew Tulloch,
  Srinivas Sridharan, Xing Liu, Mustafa Ozdal, Jade Nie, Jongsoo Park, et~al.
\newblock Software-hardware co-design for fast and scalable training of deep
  learning recommendation models.
\newblock In {\em Proceedings of the 49th Annual International Symposium on
  Computer Architecture}, pages 993--1011, 2022.

\bibitem{xiong2020layer}
Ruibin Xiong, Yunchang Yang, Di~He, Kai Zheng, Shuxin Zheng, Chen Xing,
  Huishuai Zhang, Yanyan Lan, Liwei Wang, and Tieyan Liu.
\newblock On layer normalization in the {T}ransformer architecture.
\newblock In {\em International Conference on Machine Learning}, pages
  10524--10533. PMLR, 2020.

\bibitem{zhang2019rootmeansquarelayer}
Biao Zhang and Rico Sennrich.
\newblock Root mean square layer normalization.
\newblock {\em arXiv preprint arXiv:1910.07467}, 2019.

\end{thebibliography}

\end{document}